\documentclass[runningheads]{llncs}

\usepackage{graphicx}
\usepackage{courier} % bold keywords in listing

\usepackage{booktabs}
\usepackage{makecell} % for thick table lines via Xhline
\usepackage{cite}
\usepackage{hyperref}
\usepackage{algorithm,float}
\usepackage[noend]{algpseudocode}

\usepackage{listings}
\usepackage{xcolor} % change text color
\usepackage{multirow}
\usepackage{array}

\usepackage{todonotes}
\usepackage{stmaryrd} % arrow

\usepackage{algorithm,float}
\usepackage[caption=false]{subfig}
\usepackage[export]{adjustbox} % subfloat vertical alignment (https://tex.stackexchange.com/questions/296624/subfloat-vertical-alignment-in-latex)

\lstdefinelanguage{sparql} {
language={SQL},
    alsoletter={-},
    morekeywords={skos,owl,dbr,rdf,sem,so,uner,oekg-s},
}

\newcommand{\oekg}{\textit{OEKG}}
\newcommand{\eventkg}{\textit{Event\-KG}}

\newcommand{\ekgl}{\textit{Event\-KG\textsubscript{light}}}
\newcommand{\vquanda}{\textit{VQuAnDa}}
\newcommand{\ekgclick}{\textit{Event\-KG\-+Click}}
\newcommand{\mlm}{\textit{MLM}}
\newcommand{\uner}{\textit{UNER}}
\newcommand{\infospread}{\textit{InfoSpread}}
\newcommand{\dtime}{\textit{TIME}}

\newcommand{\voc}[2]{\texttt{#1:\allowbreak #2}}
\newcommand\schema[1]{{\normalfont\fontfamily{cmvtt}\selectfont #1}}
\newcommand{\rev}[1]{#1}

\begin{document}
%{\let\thefootnote\relax\footnotetext{Copyright \textcopyright\ 2021 for this paper by its authors. Use permitted under Creative Commons License Attribution 4.0 International (CC BY 4.0).}}

{\let\thefootnote\relax\footnotetext{{The definitive version of this work was published in the Proceedings of the 2nd International Workshop on Cross-lingual Event-centric Open Analytics
co-located with the 30th The Web Conference (WWW 2021).}}}

\title{OEKG: The Open Event Knowledge Graph}

\author{Simon Gottschalk\inst{1} \and Endri Kacupaj\inst{2} \and Sara Abdollahi\inst{1} \and Diego Alves\inst{3} \and Gabriel Amaral\inst{4} \and Elisavet Koutsiana\inst{4} \and Tin Kuculo\inst{1} \and Daniela Major\inst{5} \and Caio Mello\inst{5} \and Gullal S. Cheema\inst{6} \and Abdul Sittar\inst{7} \and Swati\inst{7} \and Golsa Tahmasebzadeh\inst{6} \and Gaurish Thakkar\inst{3}}

\authorrunning{Gottschalk et al.}

\institute{
L3S Research Center, Leibniz Universität Hannover, Germany \\
\email{\{gottschalk,abdollahi,kuculo\}@L3S.de}
\and
University of Bonn, Germany \\
\email{kacupaj@cs.uni-bonn.de}
\and
University of Zagreb, Croatia \\
\email{dfvalio@ffzg.hr, gthakkar@m.ffzg.hr}
\and
King's College London, United Kingdom \\
\email{\{gabriel.maia\_rocha\_amaral,elisavet.koutsiana\}@kcl.ac.uk}
\and
School of Advanced Study, University of London, United Kingdom
\email{\{Daniela.Major,caio.mello\}@sas.ac.uk}
\and
TIB -- Leibniz Information Centre for Science and Technology, Hannover, Germany \\
\email{\{gullal.cheema,golsa.tahmasebzadeh\}@tib.eu}
\and
Jožef Stefan Institute and Jožef Stefan International Postgraduate School, Slovenia \\
\email{\{abdul.sittar,swati\}@ijs.si}
}

\maketitle

\begin{abstract}
Accessing and understanding contemporary and historical events of global impact such as the US elections and the Olympic Games is a major prerequisite for cross-lingual event analytics that investigate event causes, perception and consequences across country borders. In this paper, we present the Open Event Knowledge Graph (\oekg{}), a multilingual, event-centric, temporal knowledge graph composed of seven different data sets from multiple application domains, including question answering, entity recommendation and named entity recognition. These data sets are all integrated through an easy-to-use and robust pipeline and by linking to the event-centric knowledge graph EventKG. We describe their common schema and demonstrate the use of the \oekg{} at the example of three use cases: type-specific image retrieval, hybrid question answering over knowledge graphs and news articles, as well as language-specific event recommendation. The \oekg{} and its query endpoint are publicly available.
\end{abstract}

\section{Introduction}
\label{sec:introduction}

Contemporary and historical events such as the US presidential elections, the Olympic Games and major earthquakes change the world. Their media coverage, their varying perception by different communities, their historical evolution and potentially global impact make cross-lingual event analytics a significant research topic in various fields of studies, including social science, computer science and digital humanities \cite{rogers2013digital, time}.

When performing cross-lingual event analytics, the requirements towards event knowledge representation are manifold, given the heterogeneity, dynamicity and multilingualism of events \cite{gottschalk2018towards}. Until now, there exists a large variety of event-related data sets \cite{gottschalk2019eventkg,eventkgclick,time,sittardataset} that may help understand specific characteristics of events, but they are barely connected by now. This calls for new models and processes that enable intuitive access to the event-related knowledge spread across the world.

In this paper, we present the \oekg{}, the Open Event Knowledge Graph, which makes a step towards a holistic representation of event knowledge by the integration of event-related data sets from multiple and diverse application domains such as Question Answering, entity recommendation and Named Entity Recognition. Also, these data sets originate from different data collections, including knowledge graphs and news articles. 
\rev{One of these knowledge graphs is \ekgl{}, a new version of the event-centric and multilingual knowledge graph \eventkg{}~\cite{gottschalk2019eventkg}. The \oekg{} is built on top of \ekgl{}, allowing for easier integration of additional data sets using RDF named graphs. We propose an efficient and robust pipeline facilitating this integration of several data sets in an easy-to-use manner.}

\begin{figure}[ht]
\centering
\begin{subfloat}[Example event in the \oekg{}.]{
  \frame{\includegraphics[width=.46\linewidth]{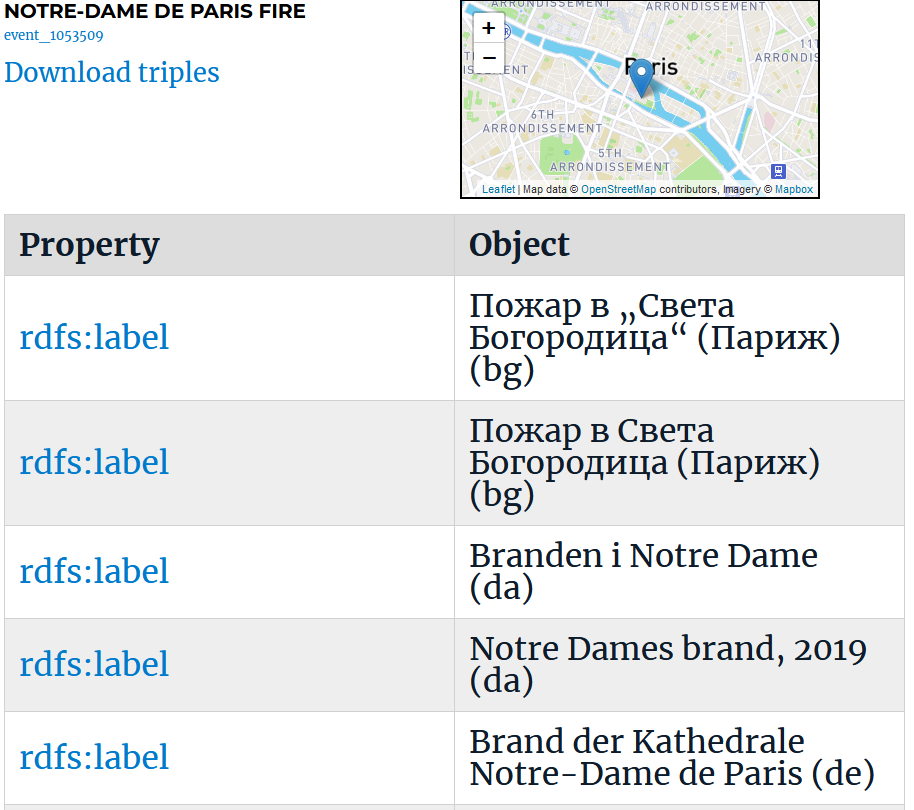}} \label{fig:intro1}
  }
\end{subfloat}%
\quad
\begin{subfloat}[Example place in the \oekg{}.]{
  \frame{\includegraphics[width=.46\linewidth]{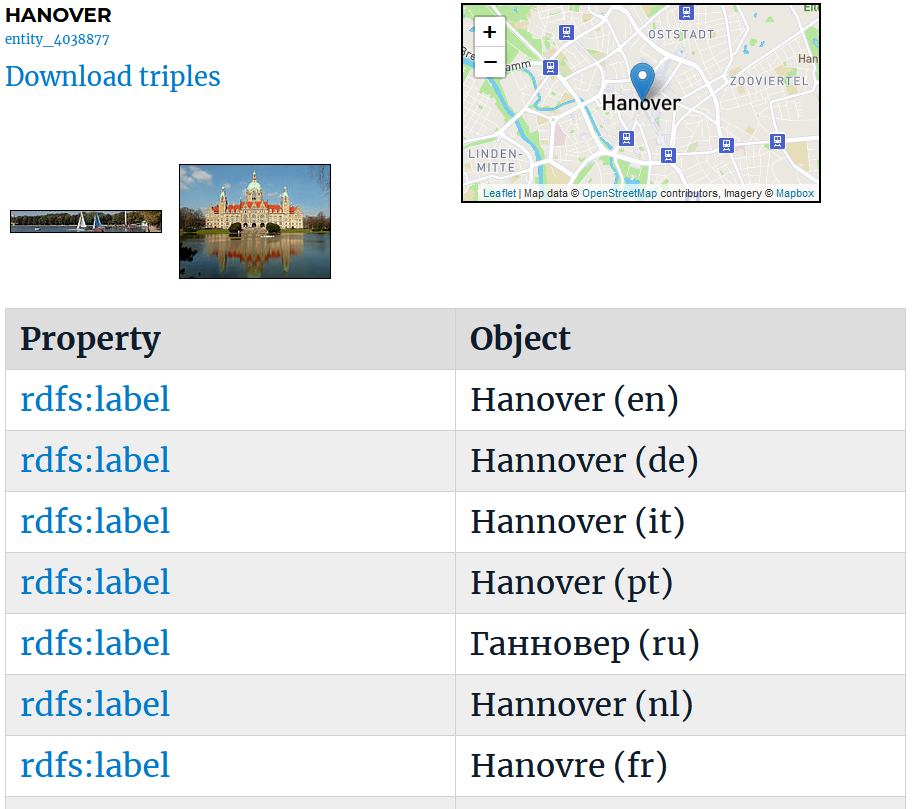}} \label{fig:intro2}
  }
\end{subfloat}

\begin{subfloat}[Example news article in the \oekg{}.]{
  \frame{\includegraphics[width=.46\linewidth]{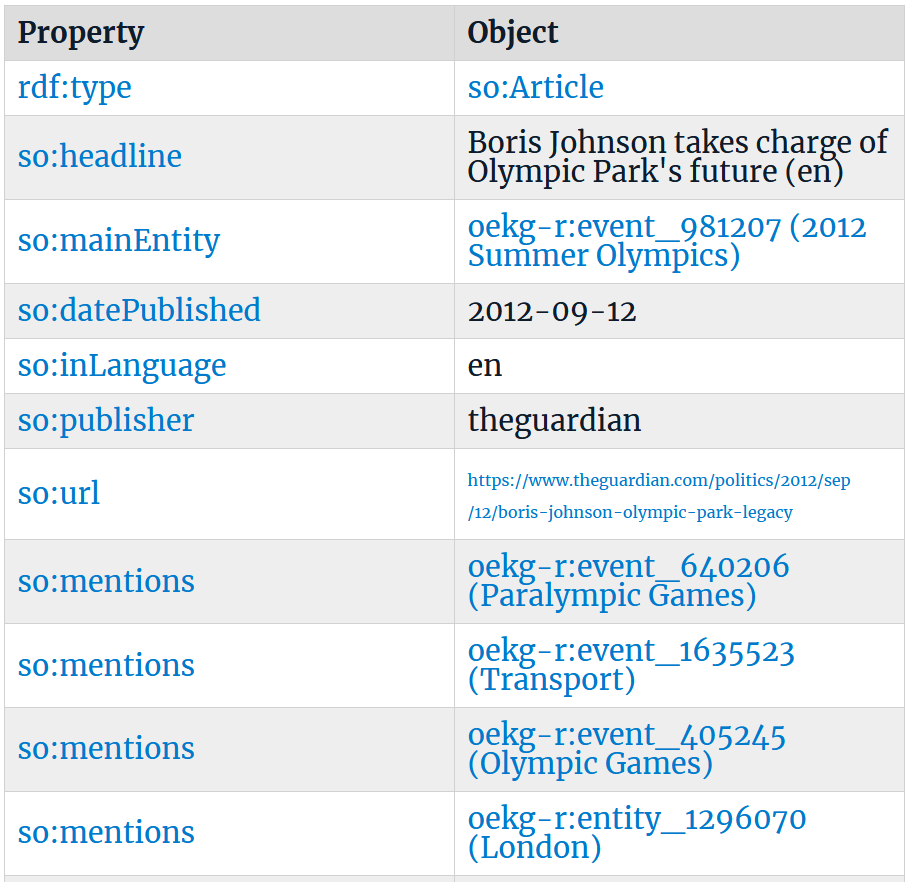}} \label{fig:intro3}
  }
\end{subfloat}%
\quad
\begin{subfloat}[Example question in the \oekg{}.]{
  \frame{\includegraphics[width=.46\linewidth]{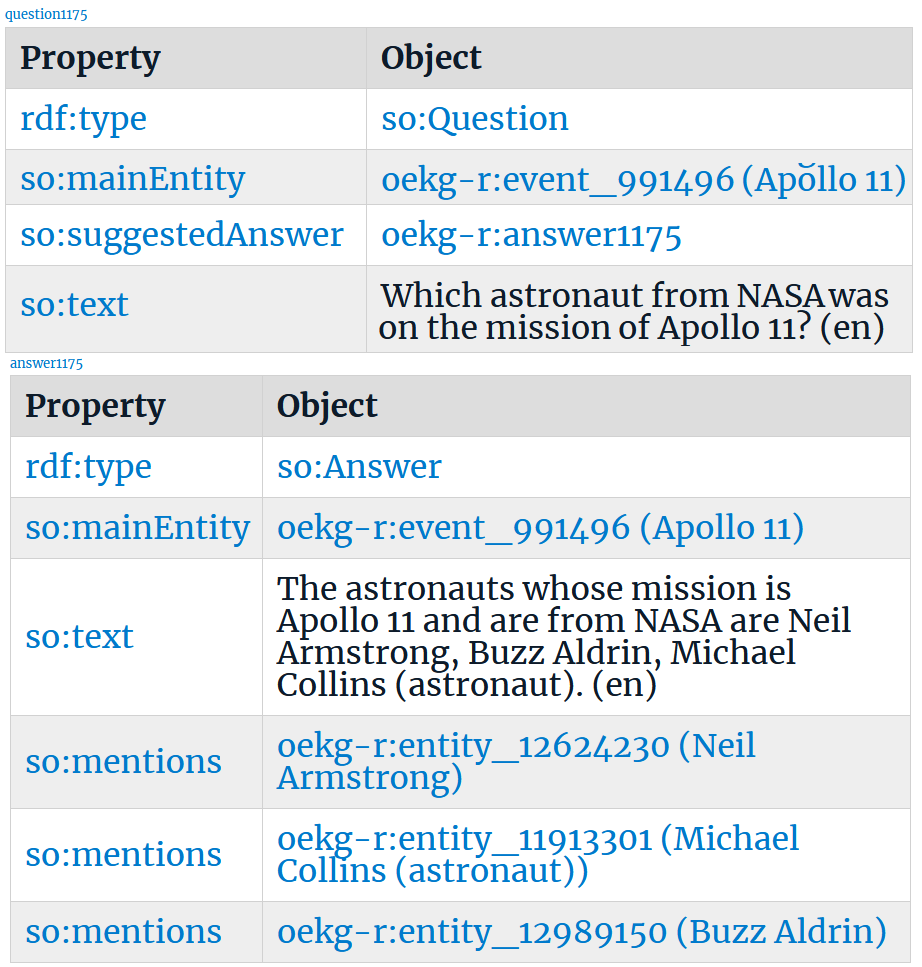}} \label{fig:intro4}
  }
\end{subfloat}

\caption[Example resources in the \oekg{}]{Example resources in the \oekg{} (not all triples are shown).\footnotemark}
\label{fig:example_resources}
\end{figure}

Fig. \ref{fig:example_resources} shows four example resources of \oekg{} and thus demonstrates its versatility resulting from the integration of several data sets:

\begin{itemize}
    \item Events (Fig. \ref{fig:intro1}): Events are at the core of the \oekg{}. For example, the fire of the Notre-Dame in Paris is covered with its locations, labels in multiple languages, related events such as ``The Notre Dame Cathedral holds its first mass since the April 15 fire'', and more event characteristics.
    \item Places (Fig. \ref{fig:intro2}): Most events happen at specific event locations which are also part of the \oekg{}. Such places do not only hold labels and coordinates, but also images and further characteristics.
    \item News articles (Fig. \ref{fig:intro3}): Events are often reported in the media \cite{dayan1994media, leban2014event}. Therefore, the \oekg{} provides access to annotated news articles. For example, the news article entitled ``Boris Johnson takes charge of Olympic Park's future'' is related to the Olympic Games.
     \item Questions and answers (Fig. \ref{fig:intro4}): Question Answering over knowledge graphs is an important natural language understanding task. The \oekg{} provides questions about events such as the Apollo 11 spaceflight, plus their answers (here, Neil Armstrong, Michael Collins and  Buzz Aldrin).
\end{itemize}

Furthermore, the \oekg{} covers several other event-related aspects, including but not limited to (temporal) event relations, language-specific relevance scores and specialised class hierarchies.  Put together, this makes the \oekg{} a versatile resource targeting a variety of potential information needs.

The \oekg{} contains more than $400$ million triples from seven data sets and is publicly available: \footnotetext{\rev{The photos of the example place are taken from Wikimedia Commons, with the second photo being licensed under the Creative Commons Attribution-Share Alike 3.0 Unported license. The maps of the example event and the example place are licensed under the Open Data Commons Open Database License (ODbL) by the OpenStreetMap Foundation (OSMF).}} We provide the triple dumps for download, a SPARQL endpoint and access to all nodes on the \oekg{} website\footnote{\url{http://oekg.l3s.uni-hannover.de}}. We also provide permanent access to the \oekg{} on Zenodo\footnote{\url{https://zenodo.org/record/4503163}}.

The remainder of this paper is organised as follows: First, we present our integration pipeline (Section \ref{sec:integration}). Then, we describe the data sets integrated into the \oekg{} (Section \ref{sec:components}) and the \oekg{} schema (Section \ref{sec:schema}). In Section \ref{sec:examples}, we provide two example use cases of the \oekg{}. Finally, we conclude in Section \ref{sec:conclusion}.

\section{Creation of the \oekg{}}
\label{sec:integration}

The creation of the \oekg{} requires an integration pipeline where a set of data sets is transformed into a single, integrated knowledge graph that provides links between all the involved resources. \ekgl{} -- a multilingual, event-centric knowledge graph later described in Section \ref{sec:components} -- serves as the base data set of the \oekg{} that contains nodes representing real-world entities and events. 

\begin{figure}[ht]
\centering
  \includegraphics[width=\linewidth]{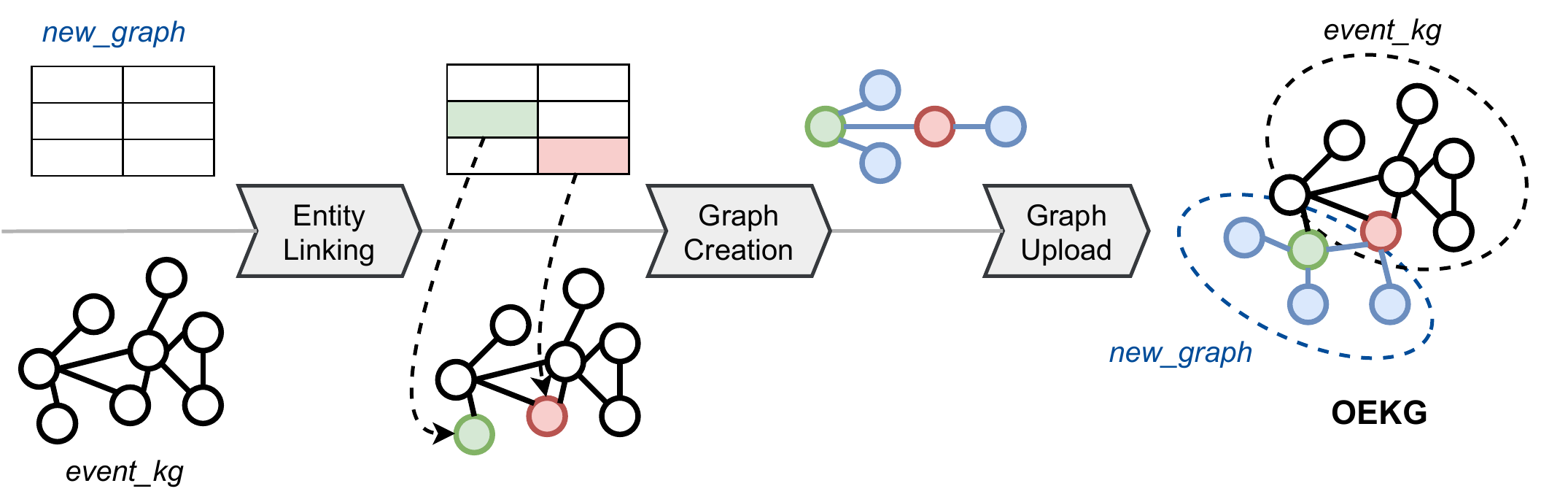}
      \caption{Example of the \oekg{} integration pipeline where a new, tabular, data set is added to the \oekg{} under the named graph \textit{new\_graph}.}
  \label{fig:pipeline}
\end{figure}

Our integration pipeline is driven by the goal to make the inclusion of a new data set into the \oekg{} as simple as possible, which allows a robust and efficient process. Only then, it is possible to integrate a large variety of data sets in an efficient and faultless way. To do so, we follow a strategy defined by Galkin et al. \cite{galkin2016integration} where the data from different sources is stored under respective named graphs. Starting from \ekgl{}, new data sets are added consecutively, each accompanied by a unique named graph. Fig. \ref{fig:pipeline} exemplifies this integration process when adding the first new data set to \ekgl{}, under the named graph \textit{new\_graph}.% In this example, the new data set is a tabular data set which does not yet hold any connection to the Linked Open Data Cloud. In a first step, some cells of the new data set are successfully linked to \ekgl{}. 

In detail, the integration process follows the following three steps: 

\begin{enumerate}
\item{Entity Linking}: We require that each graph added to \oekg{} is connected to \ekgl{}. That means any resource representing a real-world entity or event is represented by an \oekg{} resource URI. To facilitate this linking, we provide a web API that allows easy access to the \oekg{} resource URIs given Wikidata or DBpedia URIs. In our example in Fig. \ref{fig:pipeline}, some input table cells are successfully linked to \ekgl{}. 
\item{Graph Creation}: After retrieval of the \oekg{} resource URIs, a set of triples is created for each data set and serialised as an N-Triples\footnote{\url{https://www.w3.org/TR/n-triples/}} file, using the RDFLib Python library\footnote{\url{https://rdflib.dev/}}. In our example, a graph consisting of five nodes is created, two of them being already part of the \oekg{}.
\item{Graph Upload}: We provide another API method that allows uploading an N-Triples file together with the identifier of a named graph. The respective triples are then added to the \oekg{}. In our example, the resulting graph consists of two subgraphs that can be queried in isolation or together.
\end{enumerate}

\subsection{Example}

Consider Algorithm \ref{alg:integration} for an example of our integration pipeline. In this example, the new data set to be added to the \oekg{} under the named graph \textit{news} contains one news article about Barack Obama. First, the \oekg{} URI of Barack Obama is retrieved via the provided API method using the English Wikipedia label (line \ref{alg:line:el}). Second, a graph is created consisting of two triples and serialised into an RDF file (lines \ref{alg:line:triples1} - \ref{alg:line:triples2})\footnote{Relevant prefixes used by the \oekg{} are later defined in Table \ref{tab:prefixes}.}. Third, this file is uploaded via the provided API method (line \ref{alg:line:res}). In this example, one new node is added to the \oekg{} \rev{(\voc{oekg-r}{articleId})  connected to an existing node (\voc{oekg-r}{entityId})}.

\begin{algorithm}[t]
\caption{Example: Extension of the \oekg{} with a data set \textit{news} that has an article about Barack Obama}\label{alg:integration}
\begin{algorithmic}[1]
\Procedure{ExtendOEKG}{$e$}

\State graphName $\gets$ "news" .

\Comment{Entity Linking}
\State entityId $\gets$ getId$($"en"$,$"Barack\_Obama"$)$\label{alg:line:el}

\Comment{Graph Creation}
\State $G \gets$ new Graph$($graphName$)$\label{alg:line:triples1}
\State articleId $\gets$ "article1"
\State $G$.add$($oekg-r:articleId, rdf:type, so:Article$)$
\State $G$.add$($oekg-r:articleId, so:mentions, oekg-r:entityId$)$
\State fileName $\gets$ storeGraphIntoFile$(G)$\label{alg:line:triples2}

\Comment{Graph Upload}
\State uploadGraph$($fileName$,$ graphName$)$\label{alg:line:res}

\EndProcedure
\end{algorithmic}
\end{algorithm}

\subsection{Schema Extension}
\label{subsec:schema_generation}

If possible, the data sets were transformed into triples using the \ekgl{} schema of the base graph. Otherwise, the use of standard vocabularies such as schema.org\footnote{\url{https://schema.org/}} was encouraged. In every other case, schema extensions were uploaded into the \oekg{} through separate schema files using the same procedure. We will present the resulting \oekg{} schema in Section \ref{sec:schema}.

\section{Data Sets}
\label{sec:components}

The \oekg{} integrates seven data sets which are described in this section. Table \ref{tab:stats} provides an overview of these data sets, including the number of triples in the \oekg{} within their respective named graph. While some of these data sets are implicitly related to events, others add to the event knowledge from a different perspective, which will also prove useful as we will later show at the example of three use cases. 

\begin{table}[ht]
\center
\footnotesize
\caption{Statistics of the different data sets contained in the \oekg{}.}
\label{tab:stats}
\begin{tabular}{lp{8cm}l} \toprule
\textbf{Data Set} & \textbf{Short Description} & \textbf{Triples} \\ \toprule
\ekgl{} \cite{gottschalk2019eventkg} & A light-weight version of EventKG, a multilingual, event-centric, knowledge graph. & $434,752,387$  \\ \midrule
EventKG+Click \cite{eventkgclick} & A data set of language-specific event-centric user interaction traces & $118,662$ \\ \midrule
\vquanda{} \cite{kacupaj2020vquanda} & A verbalization question answering dataset & $38,243$ \\ \midrule
\mlm{} \cite{armitage2020mlm} & A benchmark dataset for multitask learning with multiple languages and modalities & $942,753$  \\ \midrule
%\begin{tabular}[c]{@{}l@{}}Information\\ Spreading \cite{sittardataset}\end{tabular} & A data set for information spreading over the news & $277,992$  \\ \midrule
\begin{tabular}[c]{@{}l@{}}InfoSpread \cite{sittardataset}\end{tabular} & A data set for information spreading over the news & $277,992$  \\ \midrule
\dtime{} \cite{time} & Two collections of news articles related to the Olympic legacy and Euroscepticism & $70,754$  \\ \midrule
\uner{} \cite{alves2020uner} & The universal named-entity recognition framework & $206,622$ \\ \Xhline{3\arrayrulewidth}
\oekg{} & The Open Event Knowledge Graph & $436,407,413$  \\ \bottomrule
\end{tabular}
\end{table}

\begin{itemize}
    \item \textbf{\ekgl{}} \cite{gottschalk2019eventkg}: The \eventkg{} is a multilingual resource incorporating event-centric information extracted from several large-scale knowledge graphs such as Wikidata, DBpedia and YAGO, as well as less structured sources such as the Wikipedia Current Events Portal and Wikipedia event lists in 15 languages. It contains nodes representing real-world entities and events and (temporal) relations between them. For the \oekg{}, we have created \ekgl{}, a light-weight version of EventKG that omits provenance information denoting the origin of relations, favouring an easier integration with other data sets.
    
    In the \oekg{}, \ekgl{} serves as the base graph that other data sets are connected to. That is to establish an agreement concerning the identification of event-related real-world objects such as persons, places and events themselves.
    \item \textbf{\ekgclick{}} \cite{eventkgclick}: \ekgclick{} is a cross-lingual dataset that reflects the language-specific relevance of events and their relations and aims to provide a reference source to train and evaluate novel models for event-centric cross-lingual user interaction. It directly builds upon \eventkg{} and language-specific information on user interactions with events, entities, and their relations derived from the Wikipedia clickstream.
    
    In the \oekg{}, \ekgclick{} can be used for recommending events to users based on actual user interaction traces. \rev{Examples of particular relevant events from a language-specific view include the 2016 Berlin truck attack from the German perspective and the 2009 Russian Premier League from the Russian perspective \cite{eventkgclick}.}
    \item \textbf{\vquanda{}} \cite{kacupaj2020vquanda}: The Verbalization QUestion ANswering DAtaset is a dataset for Question Answering (QA) over knowledge graphs that includes the verbalization of each answer. Through this verbalisation, \vquanda{} intends to completely hide any semantic technologies and provides a fluent experience between the users and the knowledge graph.
    \vquanda{} consists of $5,000$ questions accompanied by SPARQL queries and DBpedia entity links.
    
    QA over Knowledge Graphs is a common task in natural language processing \cite{dimitrakis2019survey}. Via the integration of question/answer pairs into the \oekg{}, both the question/answers pairs and the background knowledge are encapsulated into the same resource, enabling seamless training and application of QA systems.
    
    \item \textbf{\mlm{}} \cite{armitage2020mlm}: The Multiple Languages and Modalities data set is a resource for training and evaluating multitask systems in multiple modalities, for example, cross-modal (text/image) retrieval and location estimation. \mlm{} comprises text in three languages, images and location data, extracted from the Wikidata entries of $236,000$ human settlements.
    
    \mlm{} is added to the \oekg{} for adding images as an additional modality to the knowledge graph. As locations are typical event characteristics, photos of locations are an immediate benefit to the representation of events. 
    
    \item \textbf{\infospread{}} \cite{sittardataset}: The data set for Information Spreading over the News provides news articles covering three contrasting events (Global Warming, FIFA world cups and earthquakes). Initially, the goal of this data set was to understand information spreading patterns over news articles. \infospread{} contains $7,773$ news articles related to these events in five languages.
    
    News articles are often used as a means to identify events \cite{leban2014event} and oftentimes it is the media itself that makes events known to the public \cite{dayan1994media}. Therefore, the inclusion of news articles into the \oekg{} is an important step towards coverage of event-centric data from different viewpoints.
    
    \item \textbf{\dtime{}} \cite{time}: The temporal discourse analysis applied to media articles data set is a collection of Brazilian, British and Spanish news articles covering the concept of Olympic legacy and the concept of Euroscepticism.% The goal of the creation of these news collections was to analyse the concept of Olympic legacy and the concept of Euroscepticism across languages.
    
    With the collection of news articles to specified events, the \oekg{} serves as an example for in-depth analysis of single events through knowledge graphs.
    
    \item \textbf{\uner{}} \cite{alves2020uner}: The Universal Named Entity Recognition framework proposes a 4-level class hierarchy for training and testing Named Entity Recognition tools. For example, \uner{} contains the class \texttt{Earthquake}, which is a leaf node of the following branch of superclasses: \texttt{Natural}, \texttt{Na\-tu\-ral\-Phe\-nome\-non}, \texttt{Event} and \texttt{Name}.
    
    In the \oekg{}, \uner{} adds to the already given class hierarchy from the DBpedia ontology. Given how challenging it is to recognise named events in texts \cite{marujo2012recognition}, we envision that the inclusion of \uner{} classes into the \oekg{} can help training and evaluating NER systems in the specific context of event-centric data.
\end{itemize}

Following the integration pipeline described in Section \ref{sec:integration}, the described data sets were added to the \oekg{}. For additional information or increased interlinkage with \ekgl{}, some data sets were extended before:

\begin{itemize}
    \item Via the Wikifier\footnote{\url{http://wikifier.org/}} and spaCy\footnote{\url{https://spacy.io/}}, entities and events mentioned in news articles (\dtime{} and \infospread{}) were identified. This is to establish a connection between the news articles and \ekgl{}: Given this connection, one may query for news articles about specific events or entities.
    \item \rev{Sentiment analysis, i.e., the computational study of people's opinions, sentiments, emotions, moods, and attitudes \cite{liu2020sentiment}, contributes towards the understanding of natural-language texts and can, in particular, facilitate an analysis of news articles across languages~\cite{lo2017multilingual}. In the \oekg{}, we enrich news articles by employing the sentiment detection system SentiStrength~\cite{thelwall2010sentiment} on their headlines.} That way, the \oekg{} enables queries for particularly positive or negative news articles, potentially initiating further event-centric analyses of the news articles in the context of specific events.
    \item \rev{To further increase the linkage between different sources,} the \uner{} classes were aligned to the DBpedia ontology using the \texttt{skos} vocabulary\footnote{\url{https://www.w3.org/TR/swbp-skos-core-spec/}} when possible.
\end{itemize}
\section{Schema}
\label{sec:schema}

\begin{figure}[ht]
\centering
  \includegraphics[width=\linewidth]{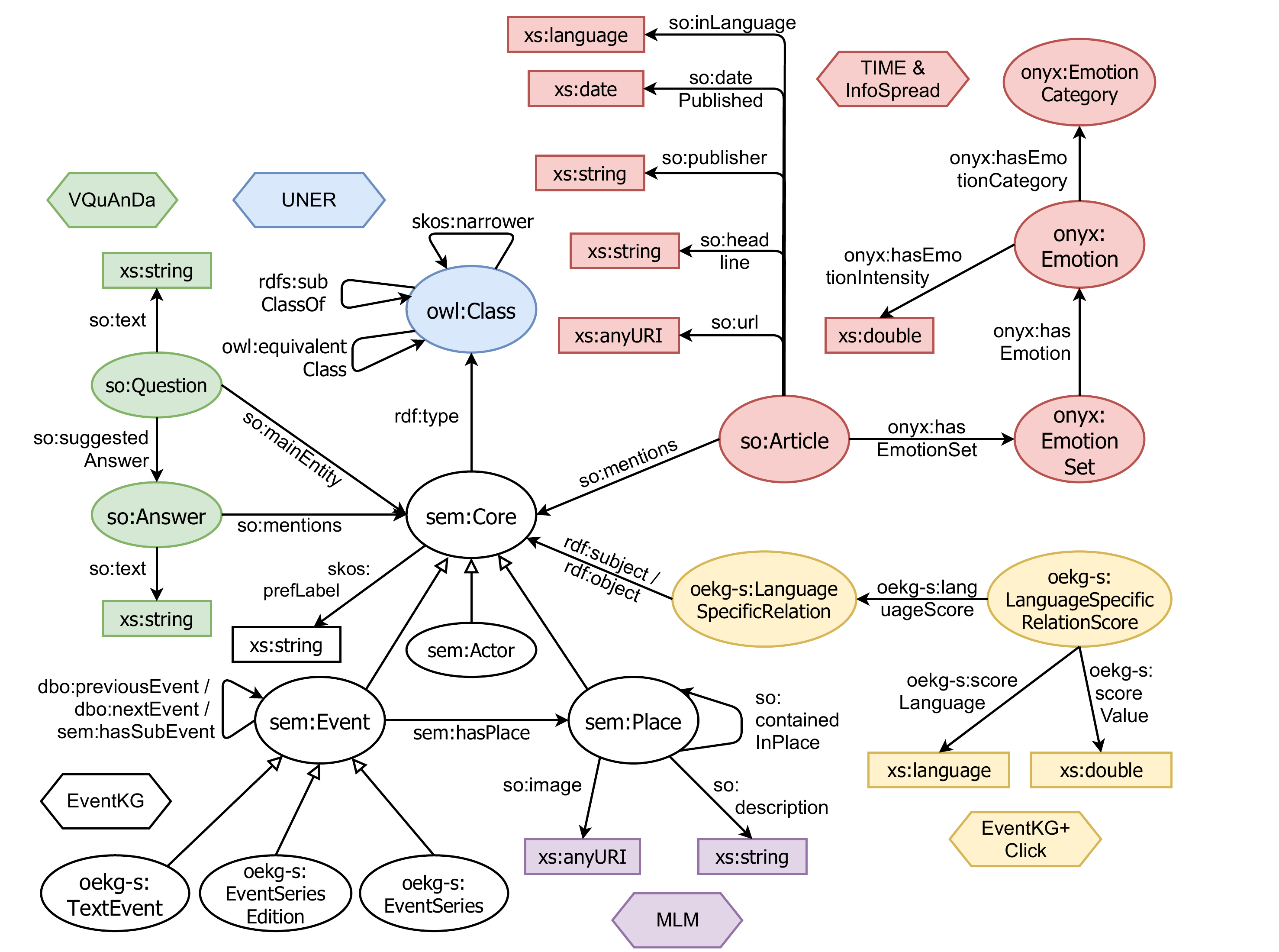}
      \caption{Excerpt of the \oekg{} schema. $\rightarrowtriangle$ marks \voc{owl}{subClassOf} relations. Regular arrows mark the \schema{rdfs:domain} and \schema{rdfs:range} restrictions on properties. Classes are coloured w.r.t. the data set for which they have been added. For brevity, we have omitted classes regarding relations between entities and events, as well as temporal attributes from the EventKG schema.}
  \label{fig:schema}
\end{figure}

Fig. \ref{fig:schema} shows the \oekg{} schema. As described in Section \ref{subsec:schema_generation}, this schema is based on the \ekgl{} schema and then extended by demand. Prefixes used in the \oekg{} schema and in the remainder of this paper are listed in Table \ref{tab:prefixes}\footnote{For a full list of prefixes used in the \oekg{}, see \url{oekg.l3s.uni-hannover.de/sparql}.}.

\begin{table}[th]
\center
\caption{Selected prefixes used by the \oekg{}.}
\label{tab:prefixes}
\begin{tabular}{@{}ll@{}}
\toprule
\textbf{Prefix} & \textbf{URI} \\ \midrule
oekg-r: & http://oekg.l3s.uni-hannover.de/resource/ \\
oekg-s: & http://oekg.l3s.uni-hannover.de/schema/ \\
oekg-g: & http://oekg.l3s.uni-hannover.de/graph/ \\
uner: & http://oekg.l3s.uni-hannover.de/uner/ \\
so: & http://schema.org/ \\
rdf: & http://www.w3.org/1999/02/22-rdf-syntax-ns\# \\
rdfs: & http://www.w3.org/2000/01/rdf-schema\# \\
xs: & http://www.w3.org/2001/XMLSchema\# \\
sem: & http://semanticweb.cs.vu.nl/2009/11/sem/ \\
onyx: & http://www.gsi.dit.upm.es/ontologies/onyx/ns\# \\
skos: & http://www.w3.org/2004/02/skos/core\# \\ \bottomrule
\end{tabular}
\end{table}

In detail, the different data sets contribute to the following parts of the \oekg{} schema:

\begin{itemize}
    \item \textbf{\ekgl{}}: The \eventkg{} schema is based on the Simple Event Model (\schema{sem})\footnote{\url{https://semanticweb.cs.vu.nl/2009/11/sem/}} and its three main classes \voc{sem}{Event}, \voc{sem}{Actor} and \voc{sem}{Place}, that are connected via \voc{sem}{hasPlace} and (temporal) relations modeled by \voc{oekg-s}{Relation} (omitted from Fig. \ref{fig:schema} for brevity). \eventkg{} further distinguishes between different types of events (\voc{oekg-s}{Text\-Event}, \voc{oekg-s}{Event\-Series} and \voc{oekg-s}{Event\-Se\-ries\-Edit\-ion}). In comparison to the \eventkg{} schema, \ekgl{} omits link count relations and adds the \voc{skos}{pref\-Label} to entities for a more efficient access to their labels.
    \item \textbf{\ekgclick{}}: \rev{To model language-specific, weighted relations for the representation of event-centric cross-lingual user interaction, we have introduced two new classes:} \voc{oekg-s}{Lan\-guage\-Speci\-fic\-Re\-lat\-ion} that assigns one or more instances of \voc{oekg-s}{Lan\-guage\-Speci\-fic\-Re\-lat\-ion\-Score} to a source entity and a target entity. Such instances hold the score between the source and target entity in a specific language.
    \item \textbf{\vquanda{}}: A question, its suggested answer and their verbalisation are represented using \schema{schema.org}'s classes \voc{so}{Question} and \voc{so}{Answer}. Entities that appear in the question text are linked to \ekgl{} instances via \voc{so}{main\-Entity}, entities in the answer via \voc{so}{mentions}.
    \item \textbf{\mlm{}}: Images are assigned to places via \voc{so}{image}, descriptions via \voc{so}{description}.
    \item \textbf{\infospread{} and \dtime{}}: News articles are represented via \voc{so}{Article} and the respective properties denoting the headline (\voc{so}{headline}), for instance. News articles are connected to \ekgl{} instances via \voc{so}{mentions}, which denote the appearance of an \oekg{} entity or event in the text. For the representation of news articles' sentiment, we follow the schema of the TweetsKB \cite{fafalios2018tweetskb}, using the \schema{onyx} vocabulary and its classes \voc{onyx}{Emotion\-Set}, \voc{onyx}{Emot\-ion} and \voc{onyx}{Emotion\-Cate\-gory} to assign a set of emotions of different strengths to a news article.
    \item \textbf{\uner{}}: Entities are assigned \uner{} classes using \voc{rdf}{type}. Furthermore, the \uner{} class hierarchy and its connection to the DBpedia ontology are established using the \schema{owl} and the \schema{skos} vocabulary.
\end{itemize}
\section{Example Use Cases}
\label{sec:examples}

In this section, we demonstrate the \oekg{} and its ability to enable integrated access over multiple datasets via three example use cases.

\subsection{Image Retrieval: \ekgl{}, \mlm{} \& \uner{}}

Event classification in images is an important task for various applications in the fields of computer vision, including geolocation estimation and place classification \cite{muller2020ontology}. Such tasks typically rely on the existence of a well-defined class hierarchy and the availability of images. The \oekg{} facilitates queries both for the \uner{} type hierarchy specifically designed for Named Entity Recognition, and for images of locations, using the \mlm{} data. In combination, event locations in \ekgl{}, \mlm{}'s image links, and the \uner{} type hierarchy enable retrieval of images relevant for specific event types.

We demonstrate the \oekg{}'s potential for image retrieval by an example query for images from earthquake regions shown in Listing \ref{lst:sparql1}: It queries for entities typed as earthquakes using the \voc{uner}{Earthquake} class, their locations (\ekgl{}) and the images assigned to such locations (\mlm{}). Table \ref{tab:example1} presents selected results of this query, including a photo of the port of Messina and more.

%They are licensed under the Creative Commons Attribution-Share Alike 3.0 Unported, 2.5 Generic, 2.0 Generic and 1.0 Generic license.}.

\begin{lstlisting}[float=ht,captionpos=b, caption=SPARQL query: Images of locations where earthquakes happened., label=lst:sparql1, frame=single]
SELECT DISTINCT ?Location ?Image WHERE {
  ?earthquake rdf:type uner:Earthquake ;
    sem:hasPlace ?Location  .
  ?Location so:image ?Image .
}
\end{lstlisting}

\begin{table}[ht]
\center
\caption[Selected \oekg{} results]{Selected \oekg{} results of the SPARQL query in Listing \ref{lst:sparql1}.\footnotemark}
\label{tab:example1}
\def\arraystretch{1.5}
\setlength\tabcolsep{5pt}
\begin{tabular}{m{1.4cm}|m{3.1cm}|m{3.1cm}|m{3.1cm}}
\textbf{Location} & Ferrara & Messina  & Guaranda \\ 
\textbf{Image} & \includegraphics[width=1\linewidth]{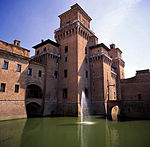} & \includegraphics[width=1\linewidth]{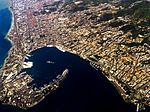} & \includegraphics[width=1\linewidth]{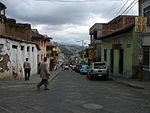} \\ 
\end{tabular}
\end{table}

\subsection{Question Answering over News Articles: \ekgl{}, \vquanda{}, \infospread{} \& \dtime{}}

Question Answering (QA) is the task of supplying precise answers to questions\footnotetext{\rev{These photos are taken from Wikimedia Commons. They are licensed under the following licenses. Ferrara: Creative Commons Attribution 2.5 Italy license. Messina: Creative Commons Attribution-Share Alike 3.0 Unported, 2.5 Generic, 2.0 Generic and 1.0 Generic. Guaranda: Creative Commons Attribution 2.0 Generic license.}}, posed by users in natural language, and is typically divided into QA over free text and QA over knowledge graphs \cite{dimitrakis2019survey}. Through the integration of \ekgl{}, \vquanda{}, \dtime{} and \infospread{} into the \oekg{}, the \oekg{} facilitates a combination of these two tasks, i.e., hybrid approaches: We can query for news articles which specifically mention the entities part of the question/answer pair. This way, two sources for answering the question can be provided: the \oekg{} itself, as well as the news article potentially holding the answer to the initially posed question.

For example, the query in Listing \ref{lst:sparql2} asks for a question in \vquanda{} (\schema{?question}) that is about an event (\texttt{?questionEntity rdf:type sem:Event}). The query then searches for news articles (\schema{?article}) mentioning both that event and one of the suggested answer entities. It returns the question ``Whose wife is a presenter at WWE? (en)'' and its verbalised answer ``The people whose partners are presenters at WWE are John Cena, Dwayne Johnson.'' together with the Spanish news articles entitled ``¿Qué luchador tiene el mayor porcentaje de victorias en la historia de WWE?'' (\textit{Which wrestler has the highest percentage of victories in in the history of WWE?}). The question entity ``WCE (en)'' is mentioned in the news article, as well as both answers: John Cena and Dwayne Johnson.

\begin{lstlisting}[float=ht,captionpos=b, caption=SPARQL query: News articles that mention entities of a question/answer pair., label=lst:sparql2, frame=single]
SELECT DISTINCT ?questionText ?answerText ?headline
                ?questionEntity ?answerEntity WHERE {

  ?question so:suggestedAnswer ?answer;
    so:mainEntity ?questionEntity ;
    so:text ?questionText .
  ?questionEntity rdf:type sem:Event .

  ?answer so:mentions ?answerEntity ;
    so:text ?answerText .

  ?article rdf:type so:Article ;
    so:mentions ?questionEntity, ?answerEntity ;
    so:headline ?headline .
}
\end{lstlisting}

\subsection{Event Recommendation: \ekgl{} \& \ekgclick{}}

As defined by Ni et. al, entity recommendation is the problem of suggesting a contextually-relevant list of entities in a particular context \cite{ni2020layered}. This task is particularly relevant in Web search. With the \oekg{}, we can specifically create language-specific recommendations for events and further enrich them with relevant event characteristics.

\begin{lstlisting}[float=ht,captionpos=b, caption=SPARQL query: Events related to the First World War from a Russian point of view., label=lst:sparql3, frame=single]
SELECT ?Label ?StartDate WHERE {
  ?event owl:sameAs dbr:World_War_I.
  ?r oekg-s:source ?event ;
    oekg-s:target ?target ;
    oekg-s:hasLanguageSpecificRelationScore [
      oekg-s:scoreValue ?value ;
      oekg-s:scoreLanguage 'ru'^^xsd:language 
    ] .
  ?target skos:prefLabel ?Label ;
    sem:hasBeginTimeStamp ?StartDate .
  FILTER(?value >= 0.8) .
}
ORDER BY ?StartDate
\end{lstlisting}

The query in Listing \ref{lst:sparql3} asks for events relevant to the First World War, from the Russian point of view. We filter for the most relevant related events (\texttt{FILTER(?value >= 0.8)}) and retrieve \ekgl{}'s event characteristics to order the resulting list of events chronologically. Table \ref{tab:example3} lists the results of this query, that clearly show a Russian focus. This result could be used for creating a language-specific event timeline similar to the link-based EventKG+TL system \cite{gottschalk2018eventkgtl}, but now inferred from actual user interaction traces in \ekgclick{}.

\begin{table}[ht]
\center
\caption{All OEKG results for the SPARQL query in Listing \ref{lst:sparql3}.}
\label{tab:example3}
\begin{tabular}{ll}
\multicolumn{1}{c}{\textbf{Label}} & \multicolumn{1}{c}{\textbf{StartDate}} \\ \toprule
Brusilov Offensive \textcolor{gray}{(en)} & 1916-05-22 \\ \midrule
Russian Civil War \textcolor{gray}{(en)} & 1917-11-07 \\ \midrule
Treaty of Brest-Litovsk \textcolor{gray}{(en)} & 1918-03-03 \\ \bottomrule
\end{tabular}
\end{table}
\section{Conclusion}
\label{sec:conclusion}

In this paper, we have introduced the \oekg{} -- the Open Event Knowledge Graph\footnote{\url{http://oekg.l3s.uni-hannover.de/}}. The \oekg{} comprises event-related knowledge from seven data sets of various application domains. We have presented an easy-to-use, efficient and robust pipeline that facilitated a seamless integration of seven data sets into the \oekg{}. At the examples of image retrieval, question answering over text and event recommendation, we have exemplified three use cases of the \oekg{}.

\subsubsection*{Acknowledgements} 
% This work was partially funded by H2020-MSCA-ITN-2018-812997 under ``Cleopatra''.
The project leading to this publication has received funding from the European Union’s Horizon 2020 research and innovation programme under the Marie Skłodowska-Curie grant agreement No.~812997 (Cleopatra).

\bibliographystyle{splncs04}
\bibliography{references}

\end{document}